%% file: main_cr.tex
\documentclass[11pt,a4paper]{article}
\usepackage[hyperref]{acl2017}
\usepackage{times}
\usepackage{latexsym}
\usepackage{amsmath}
\usepackage{multirow}
\usepackage{url}
\usepackage{graphicx}
\usepackage{amssymb}
\usepackage{color}
\usepackage{lipsum} 
\usepackage{enumitem}
\setlist{nosep} 
\aclfinalcopy 

\title{Learning to Skim Text}

\author{Adams Wei Yu\thanks{~~Most of work was done when AWY was with Google.} \\
  Carnegie Mellon University  \\
  {\tt weiyu@cs.cmu.edu} \\
  \And
  Hongrae Lee \\
  Google
   \\
  {\tt hrlee@google.com} 
  \And
  Quoc V. Le \\
  Google\\
  {\tt qvl@google.com}
  }
\date{}

\begin{document}
\maketitle
\begin{abstract}
Recurrent Neural Networks are showing much promise in many sub-areas of natural language processing, ranging from document classification to machine translation to automatic question answering. Despite their promise, many recurrent models have to read the whole text word by word, making it slow to handle long documents. For example, it is difficult to use a recurrent network to read a book and answer questions about it. In this paper, we present an approach of reading text while skipping irrelevant information if needed. The underlying model is a recurrent network that learns how far to jump after reading a few words of the input text. We employ a standard policy gradient method to train the model to make discrete jumping decisions. In our benchmarks on four different tasks, including number prediction, sentiment analysis, news article classification and automatic Q\&A, our proposed model, a modified LSTM with jumping, is up to 6 times faster than the standard sequential LSTM, while maintaining the same or even better accuracy. 
\end{abstract}

\input{intro}
\input{method}
\input{experiment}

\input{rw}

\input{conclusion}

\section*{Acknowledgments}
The authors would like to thank the Google Brain Team, especially Zhifeng Chen and Yuan Yu for helpful discussion about the implementation of this model on Tensorflow. The first author also wants to thank Chen Liang, Hanxiao Liu, Yingtao Tian, Fish Tung, Chiyuan Zhang and Yu Zhang for their help during the project. Finally, the authors appreciate the invaluable feedback from anonymous reviewers.

\bibliography{acl2017}
\bibliographystyle{acl_natbib}

\end{document}

%% file: intro.tex
\section{Introduction}\label{sec:intro}
The last few years have seen much success of applying neural networks
to many important applications in natural language processing, e.g.,
part-of-speech tagging, chunking, named entity
recognition~\cite{collobert2011natural}, sentiment
analysis~\cite{socher2011semi,socher2013recursive}, document
classification~\cite{kim2014convolutional,le2014distributed,zhang2015character,dai2015semi},
machine
translation~\cite{kalchbrenner2013recurrent,sutskever2014sequence,bahdanau2014neural,sennrich2015neural,wu2016google},
conversational/dialogue
modeling~\cite{sordoni2015neural,vinyals2015neural,shang2015neural},
document summarization~\cite{rush2015neural,nallapati2016abstractive},
parsing~\cite{andor2016globally} and automatic
question answering (Q\&A)~\cite{weston2015towards,hermann2015teaching,wang2016machine,wang2016multi,trischler2016parallel,lee2016learning,seo2016bidirectional,xiong2016dynamic}. An
important characteristic of all these models is that they read all the text
available to them. While it is essential for certain applications,
such as machine translation, this characteristic also makes it slow to
apply these models to scenarios that have long input text, such as
document classification or automatic Q\&A.
However, the fact that texts are usually written with redundancy inspires us
to think about the possibility of reading selectively.

\begin{figure*}[t]
  \centering
  \includegraphics[width=\textwidth]{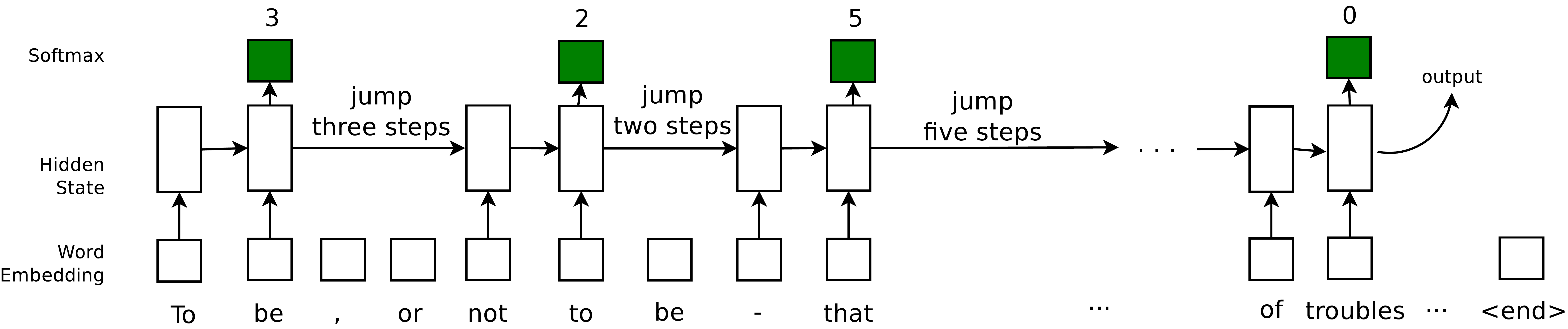}
  \caption{A synthetic example of the proposed model to process a text document. In this example, the maximum size of jump $K$ is 5, the number of tokens read before a jump $R$ is 2 and the number of jumps allowed $N$ is 10. The
    green softmax are for jumping predictions. The processing stops if a) the jumping softmax predicts a 0 or b) the jump times exceeds $N$ or c) the network processed the last token. We only show the case a) in this figure.
  }
  \label{fig:arch}
\end{figure*}

In this paper, we consider the problem of understanding documents with partial reading, and propose a modification to the basic neural architectures that
allows them to read input text with skipping. The main benefit of
this approach is faster inference because it skips irrelevant
information. An unexpected benefit of this approach is that it also
helps the models generalize better.

In our approach, the model is a recurrent network, which learns to
predict the number of jumping steps after it reads one or several input tokens. Such a discrete
model is therefore not fully differentiable, but it can be trained by
a standard policy gradient algorithm, where the reward can be
the accuracy or its proxy during training.

In our experiments, we use the basic LSTM recurrent networks~\cite{hochreiter1997long} as the
base model and benchmark the proposed algorithm on a range of document
classification or reading comprehension tasks, using various datasets such as Rotten Tomatoes~\cite{pang2005seeing},
IMDB~\cite{maas2011learning}, AG News~\cite{zhang2015character} and Children's Book Test~\cite{HillBCW15}. We find that the proposed approach
of selective reading speeds up the base model by two to six times.
Surprisingly, 
we also observe our model beats the standard LSTM in terms of accuracy.

In summary, the main contribution of our work is to design an
architecture that learns to skim text and show
that it is both faster and more accurate in practical applications of text processing.
Our model is simple and flexible enough that we anticipate it would be able to incorporate to recurrent nets with more sophisticated structures to achieve even better performance in the future. 

%% file: method.tex
\section{Methodology}\label{sec:model}

In this section, we introduce the proposed model named LSTM-Jump. We first describe its main structure, followed by the difficulty of estimating part of the model parameters because of non-differentiability. To address this issue, we appeal to a reinforcement learning formulation and adopt a policy gradient method. 

\subsection{Model Overview}

The main architecture of the proposed model is shown in
Figure~\ref{fig:arch}, which is based on an LSTM recurrent neural network. 
Before training, the number of jumps allowed $N$, the number of tokens read between every two jumps $R$ and the maximum size of
jumping $K$ are chosen ahead of time. While $K$ is a fixed parameter of the model, $N$ and $R$ are hyperparameters that can vary between training and testing. 
Also, throughout the paper, we would use $d_{1:p}$ to denote a sequence $d_1,d_2,...,d_p$.

In the following, we describe in detail how the model operates when processing text.
Given a training example $x_{1:T}$, the recurrent
network will read the embedding of the first $R$ tokens $x_{1:R}$ and output the
hidden state. Then this state is used to compute the jumping softmax that determines a distribution over the
jumping steps between $1$ and $K$. The model then samples from this
distribution a jumping step, which is used to decide the next token
to be read into the model. Let $\kappa$ be the sampled value, then the next starting token is $x_{R+\kappa}$. Such process continues until either 
\begin{enumerate}\setlength\itemsep{0em}
\item[a)] the jump softmax samples a 0; or
\item[b)] the number of jumps exceeds $N$; or
\item[c)] the model reaches the last token $x_T$.
\end{enumerate}
After stopping, as the output, the latest hidden state is further used for predicting desired targets. 
How to leverage the hidden state depends on the specifics of the task at hand. For example, for classification problems in Section~\ref{sec:np},~\ref{sec:sa} and~\ref{sec:nc}, it is directly applied to produce a softmax for classification, while in automatic Q\&A problem of Section~\ref{sec:qa}, it is used to compute the correlation with the candidate answers in order to select the best one.
Figure~\ref{fig:arch} gives an example with $K=5$, $R=2$ and $N=10$ terminating on condition a).



\subsection{Training with REINFORCE}
Our goal for training is to estimate the parameters of LSTM and possibly word embedding, which are denoted as $\theta_m$, together with the jumping action parameters $\theta_a$. Once obtained, they can be used for inference. 

The estimation of $\theta_m$ is straightforward in the tasks that can be reduced as classification problems (which is essentially what our experiments cover), as the cross entropy objective $J_1(\theta_m)$ is differentiable over $\theta_m$ that we can directly apply backpropagation to minimize. 

However, the nature of discrete jumping decisions made at every step makes it
difficult to estimate $\theta_a$, as cross entropy is no longer differentiable over $\theta_a$. Therefore, we formulate it as a reinforcement learning problem and apply
policy gradient method to train the model. 
Specifically, 
we need to maximize a reward function over $\theta_a$ which can be constructed as follows. 

Let $j_{1:N}$ be the jumping action sequence during the training with
an example $x_{1:T}$.  Suppose $h_i$ is a hidden state of the LSTM
right before the $i$-th jump $j_i$,\footnote{The $i$-th jumping step
  is usually \textit{not} $x_i$. } then it is a function of
$j_{1:i-1}$ and thus can be denoted as $h_i(j_{1:i-1})$. Now the jump
is attained by sampling from the multinomial distribution
$p(j_i|h_i(j_{1:i-1});\theta_a)$, which is determined by the jump
softmax. We can receive a reward $R$ after processing $x_{1:T}$ under the
current jumping strategy.\footnote{In the general case, one may
  receive (discounted) intermediate rewards after each jump. But in
  our case, we only consider final reward. It is equivalent to a special case that all intermediate rewards are identical and without discount.} The reward should be
positive if the output is favorable or non-positive otherwise. In our
experiments, we choose
$$R=\left\{
\begin{array}{ll}
1 & \text{if prediction correct};\\
-1 & \text{otherwise}.
\end{array}
\right.$$
Then the objective function of $\theta_a$ we want to maximize is the expected reward under the distribution defined by the current jumping policy, i.e.,
\begin{equation}\label{eqn:obj}
J_2(\theta_a) = \mathbb{E}_{p(j_{1:N};\theta_a)}[R].
\end{equation}
where 
$p(j_{1:N};\theta_a) = \prod_i p(j_{1:i}|h_i(j_{1:i-1});\theta_a).$

Optimizing this objective numerically requires computing its gradient, whose exact value is intractable to obtain as the expectation is over high dimensional interaction sequences.  By running $S$ examples,
an approximated gradient can be computed by the following REINFORCE algorithm~\cite{Williams92}:
\begin{align*}
\nabla_{\theta_a} J_2(\theta_a) 
=& \sum_{i=1}^N \mathbb{E}_{p(j_{1:N};\theta_a)}[\nabla_{\theta_a} \log p(j_{1:i}|h_i;\theta_a) R]\\
\approx& {1\over S}\sum_{s=1}^{S} 
\sum_{i=1}^N [\nabla_{\theta_a} \log p(j_{1:i}^s|h_i^s;\theta_a) R^s] 
\end{align*}
where the superscript $s$ denotes a quantity belonging to the $s$-th example. Now the term $\nabla_{\theta_a} \log p(j_{1:i}|h_i;\theta_a)$ can be computed by standard backpropagation.

Although the above estimation of $\nabla_{\theta_a} J_2(\theta_a)$ is unbiased, it may have very high variance. One widely used remedy to reduce the variance is to subtract a \textit{baseline} value $b_i^s$ from the reward $R^s$, such that the approximated gradient becomes
\begin{equation*}
\nabla_{\theta_a} J_2(\theta_a) 
\approx {1\over S}\sum_{s=1}^{S} 
\sum_{i=1}^N [\nabla_{\theta_a} \log p(j_{1:i}^s|h_i^s;\theta) (R^s - b_i^s)]
\end{equation*}
It is shown~\cite{Williams92,zaremba2015reinforcement} that any number $b_i^s$ will yield an  unbiased estimation. Here, we adopt the strategy of~\citet{mnih2014recurrent} that $b_i^s=w_bh_i^s+c_b$ and the parameter $\theta_b =\{w_b, c_b\}$ is learned by minimizing
$(R^s-b_i^s)^2$.
Now the final objective to minimize is
$$J(\theta_m,\theta_a,\theta_b) = J_1(\theta_m) - J_2(\theta_a) + \sum_{s=1}^{S}\sum_{i=1}^{N}(R^s-b_i^s)^2,$$
which is fully differentiable and can be solved by standard backpropagation.

\subsection{Inference}
During inference, we can either use sampling or greedy evaluation by
selecting the most probable jumping step suggested by the jump softmax and
follow that path. In the our experiments, we will adopt the sampling scheme.  

%% file: experiment.tex
\section{Experimental Results}\label{sec:exp}
\begin{table*}[t]
  \small
\begin{center}
\begin{tabular}{|c|c|c|c|c|c|c|c|c|}
\hline
Task & Dataset & Level   & Vocab & AvgLen& \#train& \#valid& \#test & \#class\\\hline
Number Prediction& synthetic & word & 100 & 100 words & 1M& 10K & 10K & 100\\
Sentiment Analysis & Rotten Tomatoes & word  & 18,764 & 22 words& 8,835& 1,079 & 1,030& 2\\
Sentiment Analysis & IMDB & word  & 112,540 & 241 words & 21,143 & 3,857& 25,000& 2\\
News Classification & AG & character  & 70 & 200 characters & 101,851& 18,149 & 7,600& 4\\
Q/A & Children Book Test-NE & sentence & 53,063  & 20 sentences & 108,719 & 2,000 & 2,500 & 10 \\
Q/A & Children Book Test-CN & sentence & 53,185 &  20 sentences & 120,769 & 2,000 & 2,500 & 10 \\
\hline
\end{tabular}
\caption{Task and dataset statistics.}
\label{table:dataset}
\end{center}
\end{table*}

In this section, we present our empirical studies to understand the
efficiency of the proposed model in reading text. The tasks under
experimentation are: synthetic number prediction, sentiment
analysis, news topic classification and automatic question answering.
Those, except the first one, are representative tasks in text reading involving different
sizes of datasets and various levels of text
processing, from character to word and to sentence. Table~\ref{table:dataset} summarizes the statistics of the
dataset in our experiments.

To exclude the potential impact of advanced models, we restrict our
comparison between the vanilla LSTM~\cite{hochreiter1997long} and our
model, which is referred to as LSTM-Jump.  In a nutshell, we show
that, while achieving the same or even better testing accuracy, our model is up to 6 times and 66 times
faster than the baseline LSTM model in real and synthetic datasets, respectively, as we are able to selectively
skip a large fraction of text.

In fact, the proposed model can be readily extended to other recurrent
neural networks with sophisticated mechanisms such as attention and/or
hierarchical structure to achieve higher accuracy than those presented
below. However, this is orthogonal to the main focus of this work and
would be left as an interesting future work.

\paragraph{General Experiment Settings}
We use the Adam optimizer~\cite{kingma2014adam} with a learning rate
of 0.001 in all experiments. We also apply gradient clipping to all
the trainable variables with the threshold of 1.0. 
The
dropout rate between the LSTM layers is 0.2 and the embedding dropout
rate is 0.1. We repeat the notations $N,K,R$ defined previously in Table~\ref{table:notation},
so readers can easily refer to when looking at
Tables
\ref{table:rt},\ref{table:imdb},\ref{table:ag} and \ref{table:cbt}.
While $K$ is fixed during both training and testing, we
would fix $R$ and $N$ at training but vary their values during test to see the impact of parameter changes. Note that $N$ is essentially a constraint which can be relaxed. Yet we prefer to enforce this constraint here to let the model learn to read fewer tokens. Finally, the reported test time
is measured by running one pass of the whole test set instance by instance, and the speedup is over the base LSTM model. The code is
written with
TensorFlow.\footnote{\mbox{\url{https://www.tensorflow.org/}}}

\begin{table}[h!]
\begin{center}
\begin{tabular}{|c|c|}
\hline
Notation & Meaning  \\\hline
$N$ & number of jumps allowed \\
$K$ &  maximum size of jumping  \\
$R$ & number of tokens read before a jump\\
 \hline
\end{tabular}
\caption{Notations referred to in experiments.}
\label{table:notation}
\end{center}
\end{table}

\subsection{Number Prediction with a Synthetic Dataset}\label{sec:np}
We first test whether LSTM-Jump is
indeed able to learn how to jump if a \textit{very clear} jumping
signal is given in the text. The input of the task is a sequence of
$L$ positive integers $x_{0:T-1}$
and the output is simply
$x_{x_0}$. That is, the output is chosen from the input sequence, with
index determined by $x_0$ . Here are two examples to illustrate the
idea:
\begin{align*}
\mbox{input1}: \underline{4},5,1,7, \underline{\underline{6}} ,2.  ~~\mbox{output1}: 6\\
\mbox{input2}: \underline{2},4,\underline{\underline{9}},4,5,6.  ~~\mbox{output2}: 9
\end{align*}
One can see that $x_0$ is essentially the oracle jumping signal, i.e. the indicator of how many steps
the reading should jump to get the exact output and obviously, the
remaining number of the sequence are useless. After reading the first
token, a ``smart'' network should be able to learn from the training
examples to jump to the output position, skipping the rest. 

We generate 1 million training and 10,000 validation examples with the
rule above, each with sequence length $T = 100$. We also impose $1\le
x_0<T$ to ensure the index is valid.  We find that directly training
the LSTM-Jump with full sequence is unlikely to converge, therefore,
we adopt a curriculum training scheme. More specifically, we generate
sequences with lengths $\{10, 20, 30, 40, 50, 60, 70, 80, 90, 100\}$
and train the model starting from the shortest. Whenever the training
accuracy reaches a threshold, we shift to longer sequences. We
also train an LSTM with the same curriculum
training scheme. The
training stops when the validation accuracy is larger than $98\%$.
We choose such stopping criterion simply because 
it is the highest that both models can 
achieve.\footnote{In fact, our model can get higher but we stick to $98\%$ for ease of comparison.}
All the networks are single
layered, with hidden size 512,  embedding size 32 and batch size 100. 
During testing, we generate sequences of
lengths 10, 100 and 1000 with the same rule, each having 10,000 examples.
As the training size is large enough, we do not have to worry about
overfitting so dropout is not applied. In fact, we find that the
training, validation and testing accuracies are almost the same.

The results of LSTM and our method, LSTM-Jump, are shown in Table
\ref{table:np}. The first observation is that LSTM-Jump is faster than
LSTM; the longer the sequence is, the more significant speed-up
LSTM-Jump can gain. This is because the well-trained LSTM-Jump is aware of the jumping
signal at the first token and hence can directly jump to the output
position to make prediction, while LSTM is agnostic to the signal and
has to read the whole sequence. As a result, the reading speed
of LSTM-Jump is hardly affected by the length of sequence, but that of
LSTM is linear with respect to length. Besides, LSTM-Jump also
outperforms LSTM in terms of test accuracy under all cases. This is
not surprising either, as LSTM has to read a large amount of tokens
that are potentially not helpful and could interfere with the
prediction. In summary, the results indicate LSTM-Jump is able to
learn to jump if the signal is clear.

\begin{table}[tb]
\small
\begin{center}
\begin{tabular}{|c|c|c|c|}
\hline
Seq length & LSTM-Jump  & LSTM & Speedup  \\\hline
\multicolumn{4}{|c|}{Test accuracy}  \\\hline
10&	\textbf{98\%} &	96\% & n/a \\
100&	\textbf{98\%} &	96\%  & n/a\\
1000&	\textbf{90\%} &	80\% & n/a \\\hline
\multicolumn{4}{|c|}{Test time (Avg tokens read)}  \\\hline
10	& \textbf{13.5s (2.1)}  &	18.9s (10) & 1.40x \\
100 & \textbf{13.9s (2.2)} &	120.4s (100) & 8.66x \\
1000 &	\textbf{18.9s (3.0)}&	1250s (1000) & \textbf{66.14x}\\\hline
\end{tabular}
\caption{Testing accuracy and time of synthetic number prediction problem. The jumping level is number.}
\label{table:np}
\end{center}
\end{table}
 
\subsection{Word Level Sentiment Analysis with Rotten Tomatoes and IMDB datasets}\label{sec:sa}
As LSTM-Jump has shown great speedups in the synthetic dataset, we
would like to understand whether it could carry this benefit to
real-world data, where ``jumping'' signal is not explicit. So in this
section, we conduct sentiment analysis on two movie review
datasets, both containing equal numbers of positive and negative
reviews.

The first dataset is Rotten Tomatoes,
 which contains 10,662 documents. Since there is not a standard split, we randomly select around 80\% for training, 10\% for validation, and 10\% for testing. The average and maximum lengths of the reviews are 22 and 56 words respectively, and we pad each of them to 60. We choose the pre-trained word2vec embeddings\footnote{\url{https://code.google.com/archive/p/word2vec/}}~\cite{mikolov2013distributed} 
as our fixed word embedding that we do not update this matrix during training. Both LSTM-Jump and LSTM contain 2 layers, 256 hidden units and the batch size is 100. As the amount of training data is small, we slightly augment the data by sampling a continuous 50-word sequence in each padded reviews as one training sample. During training, we enforce LSTM-Jump to read 8 tokens before a jump ($R = 8$), and the maximum skipping tokens per jump is 10 ($K = 10$), while the number of jumps allowed is 3 ($N = 3$).

The testing result is reported in Table \ref{table:rt}. In a nutshell,
LSTM-Jump is always faster than LSTM under different combinations of
$R$ and $N$. At the same time, the accuracy is on par with that of
LSTM. In particular, the combination of $(R,N)= (7, 4)$ even
achieves slightly better accuracy than LSTM while having a 1.5x
speedup.

\begin{table}[h!]
\small
\begin{center}
\begin{tabular}{|c|c|c|c|c|}
\hline
Model & $(R, N)$   & Accuracy & Time & Speedup \\\hline
\multirow{ 3}{*}{LSTM-Jump} 
& (9, 2) &  0.783 & \textbf{6.3s} & \textbf{1.98x} \\
& (8, 3) &  0.789 & 7.3s & 1.71x \\    
& (7, 4) & \textbf{0.793} & 8.1s & 1.54x \\ \hline
LSTM & n/a & 0.791 & 12.5s & 1x \\\hline
\end{tabular}
\caption{Testing time and accuracy on the Rotten Tomatoes review classification dataset. The
  maximum size of jumping $K$ is set to $10$ for all the settings. The
  jumping level is word.  }
\label{table:rt}
\end{center}
\end{table}

The second dataset is
IMDB~\cite{maas2011learning},\footnote{\url{http://ai.Stanford.edu/amaas/data/sentiment/index.html}}
which contains 25,000 training and 25,000 testing movie reviews, where
the average length of text is 240 words, much longer than that of
Rotten Tomatoes. We randomly set aside about 15\% of training data as
validation set. Both LSTM-Jump and LSTM has one layer and 128 hidden
units, and the batch size is 50. Again, we use pretrained word2vec
embeddings as initialization but they are updated during training. We
either pad a short sequence to 400 words or randomly select a 400-word
segment from a long sequence as a training example. 
During training, we set $R=20, K=40$ and $N=5$.

\begin{table}[h!]
\small
\begin{center}
\begin{tabular}{|c|c|c|c|c|}
\hline
Model & $(R, N)$   & Accuracy & Time & Speedup \\\hline
\multirow{ 5}{*}{LSTM-Jump} 
& (80, 8) &  \textbf{0.894} & 769s & 1.62x \\
& (80, 3) &  0.892 & 764s & 1.63x\\    
& (70, 3) & 0.889 & 673s & 1.85x\\ 
& (50, 2) &  0.887 & 585s & 2.12x\\ 
& (100, 1) &  0.880 & \textbf{489s} & \textbf{2.54x}\\ \hline
LSTM & n/a & 0.891 &  1243s & 1x \\\hline
\end{tabular}
\caption{Testing time and accuracy on the IMDB sentiment analysis dataset.
  The maximum size of jumping $K$ is set to $40$ for all the settings. The jumping level is word.}
\label{table:imdb}
\end{center}
\end{table}

As Table \ref{table:imdb} shows, the result exhibits a similar trend
as found in Rotten Tomatoes that LSTM-Jump is uniformly faster than
LSTM under many settings. The various $(R, N)$ combinations again
demonstrate the trade-off between efficiency and accuracy. If one cares
more about accuracy, then allowing LSTM-Jump to read and jump more
times is a good choice. Otherwise, shrinking either one would bring
a significant speedup though at the price of losing some
accuracy. Nevertheless, the configuration with the highest accuracy
still enjoys a 1.6x speedup compared to LSTM. With a slight loss of
accuracy, LSTM-Jump can be 2.5x faster .

\subsection{Character Level News Article Classification with AG dataset}\label{sec:nc}
We now present results on testing the character level jumping with a news article classification problem. The dataset contains four classes of topics (World,
Sports,
Business,
Sci/Tech) from the AG's news corpus,\footnote{\url{http://www.di.unipi.it/~gulli/AG_corpus_of_news_articles.html}} a collection of more than 1 million news articles. The data we use is the subset constructed by~\citet{zhang2015character}  for classification with character-level convolutional networks. There are 30,000 training and 1,900 testing examples for each class respectively, where 15\% of training data is set aside as validation.
The non-space alphabet under use are:
\begin{verbatim}
abcdefghijklmnopqrstuvwxyz0123456
789-,;.!?:’’’/\|_@#$%ˆ&*~`+-=<>()[]{} 
\end{verbatim}
Since the vocabulary size is small, we choose 16 as the embedding size. The initialized entries of the embedding matrix are drawn from a uniform distribution in $[-0.25, 0.25]$, which are progressively updated during training. Both LSTM-Jump and LSTM have 1 layer and 64 hidden units and the batch sizes are 20 and 100 respectively. The training sequence is again of length 400 that it is either padded from a short sequence or sampled from a long one. 
During training, we set $R=30, K=40$ and $N=5$.

The result is summarized in Table~\ref{table:ag}. 
It is interesting to see that even with skipping, LSTM-Jump is not always faster than LSTM.
This is mainly due to the fact that the embedding size and hidden layer are both much smaller than those used previously, and accordingly the processing of a token is much faster. In that case, other computation overhead such as calculating and sampling from the jump softmax might become a dominating factor of efficiency.
By this cross-task comparison, we can see that the larger the hidden unit size of recurrent neural network and the embedding are, the more speedup LSTM-Jump can gain, which is also confirmed by the task below.

\begin{table}[h!]
\small
\begin{center}
\begin{tabular}{|c|c|c|c|c|}
\hline
Model &  $(R, N)$   & Accuracy & Time & Speedup \\\hline
\multirow{5}{*}{LSTM-Jump} 
& (50, 5) &  0.854 & 102s & 0.80x \\  
& (40, 6) &  0.874 & 98.1s & 0.83x \\
& (40, 5) &  0.889 & 83.0s & 0.98x\\
& (30, 5) &  0.885 & \textbf{63.6s} & \textbf{1.28x}\\
& (30, 6) & \textbf{0.893} & 74.2s & 1.10x \\ \hline
LSTM & n/a & 0.881 & 81.7s & 1x \\\hline
\end{tabular}
\caption{Testing time and accuracy on the AG news classification dataset.  The maximum size of jumping $K$ is set to $40$ for all the settings. The jumping level is character.}
\label{table:ag}
\end{center}
\end{table}

\subsection{Sentence Level Automatic Question Answering with Children's Book Test dataset}\label{sec:qa}
The last task is automatic question answering, in which we aim to test
the sentence level skimming of LSTM-Jump. We benchmark on the data set
Children's Book Test (CBT)~\cite{HillBCW15}.\footnote{\url{
 http://www.thespermwhale.com/jaseweston/babi/CBTest.tgz}} In each
document, there are 20 contiguous sentences (context) extracted from a
children's book followed by a query sentence. A word of the query is
deleted and the task is to select the best fit for this position from
10 candidates. Originally, there are four types of tasks according to the
part of speech of the missing word, from which, we choose the most
difficult two, i.e., the name entity (NE) and common noun (CN) as our
focus, since  simple language models can already achieve human-level
performance for the other two types .

The models, LSTM or LSTM-Jump, firstly read the whole query, then the
context sentences and finally output the predicted word. While LSTM
reads everything, our jumping model would decide how many context
sentences should skip after reading one sentence. Whenever a model
finishes reading, the context and query are encoded in its hidden
state $h_{\mbox{o}}$, and the best answer from the candidate words has
the same index that maximizes the following:
$$\mbox{softmax}(CWh_{\mbox{o}})\in\mathbb{R}^{10},$$ where $C\in \mathbb{R}^{10\times
  d}$ is the word embedding matrix of the 10 candidates and
$W\in\mathbb{R}^{d\times \mbox{hidden\_size}}$ is a trainable weight
variable. Using such bilinear form to select answer basically follows
the idea of~\citet{ChenBM16}, as it is shown to have good
performance. The task is now distilled to a  classification
problem of  10 classes.

We either truncate or pad each context sentence, such that they all
have length 20. The same preprocessing is applied to the query
sentences except that the length is set as 30. For both models, the
number of layers is 2, the number of hidden units is 256 and the batch size is
32. Pretrained word2vec embeddings are again used and they are not
adjusted during training. The maximum number of context sentences
LSTM-Jump can skip per time is $K = 5$ while the number of total jumping is
limited to $N=5$. We let the model jump after reading every sentence, so $R=1$ (20 words).

The result is reported in Table \ref{table:cbt}. The performance of
LSTM-Jump is superior to LSTM in terms of both accuracy and efficiency
under all settings in our experiments. In particular, the fastest
LSTM-Jump configuration achieves a remarkable 6x speedup over LSTM,
while also having respectively 1.4\% and 4.4\% higher accuracy in
Children's Book Test - Named Entity  and Children's Book Test - Common Noun.

\begin{table}[h!]
\small
\begin{center}
\begin{tabular}{|c|c|c|c|c|}
\hline
Model &  $(R, N)$   & Accuracy & Time & Speedup \\\hline
\multicolumn{5}{|c|}{Children's Book Test - Named Entity}\\\hline
\multirow{3}{*}{LSTM-Jump} 
& (1, 5) &  \textbf{0.468} & 40.9s & 3.04x  \\ 
& (1, 3) & 0.464  & 30.3s & 4.11x\\ 
& (1, 1) & 0.452  & \textbf{19.9s} & \textbf{6.26x} \\ \hline
LSTM & n/a & 0.438& 124.5s  & 1x\\\hline
\multicolumn{5}{|c|}{Children's Book Test - Common Noun}\\\hline
\multirow{3}{*}{LSTM-Jump} 
& (1, 5) &  0.493 & 39.3s & 3.09x \\
& (1, 3) &  0.487 & 29.7s & 4.09x\\    
& (1, 1) & \textbf{0.497} & \textbf{19.8s} & \textbf{6.14x} \\ \hline
LSTM & n/a & 0.453  & 121.5s  & 1x\\\hline
\end{tabular}
\caption{Testing time and accuracy on the Children's Book Test dataset.
  The maximum size of jumping $K$ is set to $5$ for all the
  settings. The jumping level is sentence.}
\label{table:cbt}
\end{center}
\end{table}

The dominant performance of LSTM-Jump over LSTM might be interpreted
as follows. After reading the query, both LSTM and LSTM-Jump know what
the question is. However, LSTM still has to process the remaining 20
sentences and thus at the very end of the last sentence, the long
dependency between the question and output might become weak that the
prediction is hampered. On the contrary, the question can guide
LSTM-Jump on how to read selectively and stop early when the answer is clear. Therefore, when it comes to the output stage, the ``memory'' is both fresh and uncluttered that a more accurate answer is
likely to be picked. 

In the following, we show two examples of how the model reads the context given a query 
(bold
face sentences are those read by our model in the increasing order). XXXXX is the missing word we want to fill. Note that due to truncation, a few sentences might look uncompleted.

\paragraph{Example 1}
In the first example, the exact answer appears in the context multiple times, which makes the task relatively easy, as long as the reader has captured their occurrences.

\begin{enumerate}\setlength\itemsep{0em}
\item[(a)] {\it \underline{Query}:} `XXXXX!
\item[(b)] {\it \underline{Context}:} 
\item {\bf said Big Klaus, and he ran off at once to Little Klaus.} 
\item `Where did you get so much money from?' 
\item {\bf `Oh, that was from my horse-skin.}
\item I sold it yesterday evening.' 
\item `That 's certainly a good price!' 
\item said Big Klaus; and running home in great haste, he took an axe, knocked all his four 
\item {\bf `Skins!}
\item {\bf skins!}
\item Who will buy skins?' 
\item he cried through the streets. 
\item All the shoemakers and tanners came running to ask him what he wanted for them.' 
\item {\bf A bushel of money for each,' said Big Klaus.}
\item `Are you mad?' 
\item {\bf they all exclaimed.}
\item `Do you think we have money by the bushel?'
\item `Skins! 
\item skins!
\item Who will buy skins?' 
\item he cried again, and to all who asked him what they cost, he answered,' A bushel
\item `He is making game of us,' they said; and the shoemakers seized their yard measures and
\item[(c)] {\it \underline{Candidates}:} Klaus $|$ Skins $|$ game $|$ haste $|$ head $|$ home $|$ horses $|$ money $|$ price$|$ streets 
\item[(d)] {\it \underline{Answer}:} Skins
\end{enumerate}

The reading behavior might be interpreted as follows.
The model tries to search for clues, and after reading sentence 8, 
it realizes that
the most plausible answer is ``Klaus" or ``Skins", as they both appear twice. ``Skins" is more likely to be the answer as it is followed by a ``!''.
The model searches further to see if "Klaus!" is mentioned somewhere, but it only finds ``Klaus'' without ``!'' for the third time. After the last attempt at sentence 14,
it is confident about the answer and stops to output with ``Skins".

\paragraph{Example 2}
In this example, the answer is illustrated by a word ``nuisance'' that does not show up in the context at all. Hence, to answer the query, the model has to understand the meaning of both the query and context and locate the synonym of ``nuisance'', which is not merely verbatim and thus much harder than the previous example. Nevertheless, our model is still able to make a right choice while reading much fewer sentences.

\begin{enumerate}\setlength\itemsep{0em}
\item[(a)] {\it \underline{Query}:} Yes, I call XXXXX a nuisance.
\item[(b)] {\it \underline{Context}:} 
\item \textbf{But to you and me it would have looked just as it did to Cousin Myra -- a very discontented}
\item ``I'm awfully glad to see you, Cousin Myra, ''explained Frank carefully, ``and your 
\item \textbf{But Christmas is just a bore -- a regular bore.'' }
\item \textbf{That was what Uncle Edgar called things that didn't interest him, so that Frank felt pretty sure of}
\item \textbf{Nevertheless, he wondered uncomfortably what made Cousin Myra smile so queerly.}
\item \textbf{``Why, how dreadful!''}
\item she said brightly. 
\item ``I thought all boys and girls looked upon Christmas as the very best time in the year.'' 
\item ``We don't, ''said Frank gloomily. 
\item \textbf{``It's just the same old thing year in and year out.} 
\item We know just exactly what is going to happen. 
\item We even know pretty well what presents we are going to get. 
\item And Christmas Day itself is always the same. 
\item We'll get up in the morning , and our stockings will be full of things, and half of 
\item Then there 's dinner. 
\item It 's always so poky. 
\item And all the uncles and aunts come to dinner -- just the same old crowd, every year, and 
\item Aunt Desda always says, `Why, Frankie, how you have grown!' 
\item She knows I hate to be called Frankie. 
\item And after dinner they'll sit round and talk the rest of the day, and that's all. 
\item[(c)] {\it \underline{Candidates}:} 
Christmas $|$ boys $|$ day $|$ dinner $|$ half $|$ interest $|$ rest $|$ stockings $|$ things $|$ uncles 
\item[(d)] {\it \underline{Answer}:} Christmas
\end{enumerate}

The reading behavior can be interpreted as follows. After reading the query, our model realizes that the answer should be something like a nuisance. Then it starts to process the text. Once it hits sentence 3, it may begin to consider ``Christmas'' as the answer, since ``bore'' is a synonym of ``nuisance''. Yet the model is not 100\% sure, so it continues to read, very conservatively -- it does not jump for the next three sentences. After that, the model gains more confidence on the answer ``Christmas'' and it makes a large jump to see if there is something that can turn over the current hypothesis. It turns out that the last-read sentence is still talking about Christmas with a negative voice. 
Therefore, the model stops to take ``Christmas'' as the output.

%
%
%

%% file: rw.tex
\section{Related Work}
\label{sec:rw}
Closely related to our work is the idea of learning visual attention
with neural
networks~\cite{mnih2014recurrent,ba2014multiple,sermanet2014attention},
where a recurrent model is used to combine visual evidence at multiple
fixations processed by a convolutional neural network. Similar to our
approach, the model is trained end-to-end using the REINFORCE
algorithm~\cite{Williams92}. However, a major difference between those work and ours is that we have to sample from discrete jumping distribution, while they can sample from continuous distribution such as Gaussian. The difference is mainly due to the inborn characteristics of text and image. In fact, as pointed out by~\citet{mnih2014recurrent}, it was difficult to learn policies over more than 25 possible discrete locations.

This idea has recently been explored in the context of natural
language processing applications, where the main goal is to filter
irrelevant content using a small
network~\cite{DBLP:journals/corr/ChoiHLPUB16}. Perhaps the most closely
related to our work is the concurrent work on learning to reason
with reinforcement learning~\cite{shen2016reasonet}. The key
difference between our work and~\citet{shen2016reasonet} is
that they focus on early stopping after multiple pass of data to ensure accuracy whereas our method focuses on selective reading with single pass to enable fast processing.


The concept of ``hard'' attention has also been used successfully in the
context of making neural network predictions more
interpretable~\cite{lei2016rationalizing}.  The key difference between
our work and~\citet{lei2016rationalizing}'s method is that our method
optimizes for faster inference, and is more dynamic in its
jumping. Likewise is the difference between our approach and the
``soft'' attention approach by~\cite{bahdanau2014neural}. 
Recently, ~\cite{HahnK16} investigate how machine can fixate and skip words, focusing on the comparison between the behavior of machine and human, while our goal is to make reading faster. They model the probability that each single word should be read in an unsupervised way while ours directly model the probability of how many words should be skipped with supervised learning.

Our method belongs to adaptive computation of neural networks, whose idea is recently explored
by~\cite{graves2016adaptive,jernite2016}, where different amount of
computations are allocated dynamically per time step. The main
difference between our method
and~\citeauthor{graves2016adaptive,jernite2016}'s methods is that our
method can set the amount of computation to be exactly zero for many
steps, thereby achieving faster scanning over texts. Even though our
method requires policy gradient methods to train, which is a
disadvantage compared to~\cite{graves2016adaptive,jernite2016}, we do
not find training with policy gradient methods problematic in our
experiments.

At the high-level, our model can be viewed as a simplified trainable
Turing machine, where the controller can move on the
input tape. It is therefore related to the prior work on Neural Turing
Machines~\cite{graves2014neural} and especially its RL
version~\cite{zaremba2015reinforcement}. Compared
to~\cite{zaremba2015reinforcement}, the output tape in our method is
more simple and reward signals in our problems are less sparse,
which explains why our model is easy to train. It is worth noting
that Zaremba and Sutskever report difficulty in using policy
gradients to train their model.

Our method, by skipping irrelevant content, shortens the length of
recurrent networks, thereby addressing the vanishing or exploding
gradients in them~\cite{hochreiter2001gradient}. The baseline method
itself, Long Short Term Memory~\cite{hochreiter1997long}, belongs to
the same category of methods. In this category, there are several
recent methods that try to achieve the same goal, such as having
recurrent networks that operate in different
frequency~\cite{koutnik2014clockwork} or is organized in a
hierarchical fashion~\cite{chan2015listen,chung2016hierarchical}.

Lastly, we should point out that we are among the recent efforts that deploy
reinforcement learning to the field of natural language processing, some of which 
have achieved encouraging 
results in the realm of such as neural symbolic machine~\citep{LiangBLFL17},
machine reasoning~\citep{shen2016reasonet} and sequence generation~\citep{RanzatoCAZ15}.

%% file: conclusion.tex
\section{Conclusions}\label{sec:conclusion}

In this paper, we focus on learning how to skim text for fast reading. In particular, we propose a ``jumping'' model that after reading every few tokens, it decides how many tokens should be skipped by sampling from a softmax. Such jumping behavior is modeled as a discrete decision making process, which can be trained by reinforcement learning algorithm such as REINFORCE. In four different tasks with six datasets (one synthetic and five real), we test the efficiency of the proposed method on various levels of text jumping, from character to word and then to sentence. The results indicate our model is several times faster than, while the accuracy is on par with the baseline LSTM model. 


%% file: main_cr.bbl
\begin{thebibliography}{}
\expandafter\ifx\csname natexlab\endcsname\relax\def\natexlab#1{#1}\fi

\bibitem[{Andor et~al.(2016)Andor, Alberti, Weiss, Severyn, Presta, Ganchev,
  Petrov, and Collins}]{andor2016globally}
Daniel Andor, Chris Alberti, David Weiss, Aliaksei Severyn, Alessandro Presta,
  Kuzman Ganchev, Slav Petrov, and Michael Collins. 2016.
\newblock Globally normalized transition-based neural networks.
\newblock {\em arXiv preprint arXiv:1603.06042\/} .

\bibitem[{Ba et~al.(2014)Ba, Mnih, and Kavukcuoglu}]{ba2014multiple}
Jimmy Ba, Volodymyr Mnih, and Koray Kavukcuoglu. 2014.
\newblock Multiple object recognition with visual attention.
\newblock {\em arXiv preprint arXiv:1412.7755\/} .

\bibitem[{Bahdanau et~al.(2014)Bahdanau, Cho, and Bengio}]{bahdanau2014neural}
Dzmitry Bahdanau, Kyunghyun Cho, and Yoshua Bengio. 2014.
\newblock Neural machine translation by jointly learning to align and
  translate.
\newblock {\em arXiv preprint arXiv:1409.0473\/} .

\bibitem[{Chan et~al.(2015)Chan, Jaitly, Le, and Vinyals}]{chan2015listen}
William Chan, Navdeep Jaitly, Quoc~V Le, and Oriol Vinyals. 2015.
\newblock Listen, attend and spell.
\newblock {\em arXiv preprint arXiv:1508.01211\/} .

\bibitem[{Chen et~al.(2016)Chen, Bolton, and Manning}]{ChenBM16}
Danqi Chen, Jason Bolton, and Christopher~D. Manning. 2016.
\newblock A thorough examination of the cnn/daily mail reading comprehension
  task.
\newblock In {\em Proceedings of the 54th Annual Meeting of the Association for
  Computational Linguistics, {ACL} 2016, August 7-12, 2016, Berlin, Germany,
  Volume 1: Long Papers\/}.

\bibitem[{Choi et~al.(2016)Choi, Hewlett, Lacoste, Polosukhin, Uszkoreit, and
  Berant}]{DBLP:journals/corr/ChoiHLPUB16}
Eunsol Choi, Daniel Hewlett, Alexandre Lacoste, Illia Polosukhin, Jakob
  Uszkoreit, and Jonathan Berant. 2016.
\newblock Hierarchical question answering for long documents.
\newblock {\em arXiv preprint arXiv:1611.01839\/} .

\bibitem[{Chung et~al.(2016)Chung, Ahn, and Bengio}]{chung2016hierarchical}
Junyoung Chung, Sungjin Ahn, and Yoshua Bengio. 2016.
\newblock Hierarchical multiscale recurrent neural networks.
\newblock {\em arXiv preprint arXiv:1609.01704\/} .

\bibitem[{Collobert et~al.(2011)Collobert, Weston, Bottou, Karlen, Kavukcuoglu,
  and Kuksa}]{collobert2011natural}
Ronan Collobert, Jason Weston, L{\'e}on Bottou, Michael Karlen, Koray
  Kavukcuoglu, and Pavel Kuksa. 2011.
\newblock Natural language processing (almost) from scratch.
\newblock {\em Journal of Machine Learning Research\/} 12(Aug):2493--2537.

\bibitem[{Dai and Le(2015)}]{dai2015semi}
Andrew~M. Dai and Quoc~V. Le. 2015.
\newblock Semi-supervised sequence learning.
\newblock In {\em Advances in Neural Information Processing Systems\/}. pages
  3079--3087.

\bibitem[{Graves(2016)}]{graves2016adaptive}
Alex Graves. 2016.
\newblock Adaptive computation time for recurrent neural networks.
\newblock {\em arXiv preprint arXiv:1603.08983\/} .

\bibitem[{Graves et~al.(2014)Graves, Wayne, and Danihelka}]{graves2014neural}
Alex Graves, Greg Wayne, and Ivo Danihelka. 2014.
\newblock Neural turing machines.
\newblock {\em arXiv preprint arXiv:1410.5401\/} .

\bibitem[{Hahn and Keller(2016)}]{HahnK16}
Michael Hahn and Frank Keller. 2016.
\newblock Modeling human reading with neural attention.
\newblock In {\em EMNLP\/}. pages 85--95.

\bibitem[{Hermann et~al.(2015)Hermann, Kocisky, Grefenstette, Espeholt, Kay,
  Suleyman, and Blunsom}]{hermann2015teaching}
Karl~Moritz Hermann, Tomas Kocisky, Edward Grefenstette, Lasse Espeholt, Will
  Kay, Mustafa Suleyman, and Phil Blunsom. 2015.
\newblock Teaching machines to read and comprehend.
\newblock In {\em Advances in Neural Information Processing Systems\/}. pages
  1693--1701.

\bibitem[{Hill et~al.(2015)Hill, Bordes, Chopra, and Weston}]{HillBCW15}
Felix Hill, Antoine Bordes, Sumit Chopra, and Jason Weston. 2015.
\newblock The goldilocks principle: Reading children's books with explicit
  memory representations.
\newblock {\em arXiv:1511.02301\/} .

\bibitem[{Hochreiter et~al.(2001)Hochreiter, Bengio, Frasconi, and
  Schmidhuber}]{hochreiter2001gradient}
Sepp Hochreiter, Yoshua Bengio, Paolo Frasconi, and J{\"u}rgen Schmidhuber.
  2001.
\newblock Gradient flow in recurrent nets: the difficulty of learning long-term
  dependencies.
\newblock In S.~C. Kremer and J.~F. Kolen, editors, {\em A Field Guide to
  Dynamical Recurrent Neural Networks\/}, IEEE press.

\bibitem[{Hochreiter and Schmidhuber(1997)}]{hochreiter1997long}
Sepp Hochreiter and J{\"u}rgen Schmidhuber. 1997.
\newblock Long short-term memory.
\newblock {\em Neural computation\/} 9(8):1735--1780.

\bibitem[{Jernite et~al.(2016)Jernite, Grave, Joulin, and
  Mikolov}]{jernite2016}
Yacine Jernite, Edouard Grave, Armand Joulin, and Tomas Mikolov. 2016.
\newblock Variable computation in recurrent neural networks.
\newblock {\em arXiv preprint arXiv:1611.06188\/} .

\bibitem[{Kalchbrenner and Blunsom(2013)}]{kalchbrenner2013recurrent}
Nal Kalchbrenner and Phil Blunsom. 2013.
\newblock Recurrent continuous translation models.
\newblock In {\em EMNLP\/}.

\bibitem[{Kim(2014)}]{kim2014convolutional}
Yoon Kim. 2014.
\newblock Convolutional neural networks for sentence classification.
\newblock {\em arXiv preprint arXiv:1408.5882\/} .

\bibitem[{Kingma and Ba(2014)}]{kingma2014adam}
Diederik Kingma and Jimmy Ba. 2014.
\newblock Adam: A method for stochastic optimization.
\newblock {\em arXiv preprint arXiv:1412.6980\/} .

\bibitem[{Koutnik et~al.(2014)Koutnik, Greff, Gomez, and
  Schmidhuber}]{koutnik2014clockwork}
Jan Koutnik, Klaus Greff, Faustino Gomez, and Juergen Schmidhuber. 2014.
\newblock A clockwork rnn.
\newblock In {\em International Conference on Machine Learning\/}.

\bibitem[{Le and Mikolov(2014)}]{le2014distributed}
Quoc~V. Le and Tomas Mikolov. 2014.
\newblock Distributed representations of sentences and documents.
\newblock In {\em International Conference on Machine Learning (ICML)\/}.

\bibitem[{Lee et~al.(2016)Lee, Kwiatkowski, Parikh, and Das}]{lee2016learning}
Kenton Lee, Tom Kwiatkowski, Ankur Parikh, and Dipanjan Das. 2016.
\newblock Learning recurrent span representations for extractive question
  answering.
\newblock {\em arXiv preprint arXiv:1611.01436\/} .

\bibitem[{Lei et~al.(2016)Lei, Barzilay, and Jaakkola}]{lei2016rationalizing}
Tao Lei, Regina Barzilay, and Tommi Jaakkola. 2016.
\newblock Rationalizing neural predictions.
\newblock {\em arXiv preprint arXiv:1606.04155\/} .

\bibitem[{Liang et~al.(2017)Liang, Berant, Le, Forbus, and Lao}]{LiangBLFL17}
Chen Liang, Jonathan Berant, Quoc Le, Kenneth~D. Forbus, and Ni~Lao. 2017.
\newblock Neural symbolic machines: Learning semantic parsers on freebase with
  weak supervision.
\newblock In {\em Proceedings of the 55th Annual Meeting of the Association for
  Computational Linguistics, {ACL} 2017: Long Papers\/}.

\bibitem[{Maas et~al.(2011)Maas, Daly, Pham, Huang, Ng, and
  Potts}]{maas2011learning}
Andrew~L Maas, Raymond~E Daly, Peter~T. Pham, Dan Huang, Andrew~Y. Ng, and
  Christopher Potts. 2011.
\newblock Learning word vectors for sentiment analysis.
\newblock In {\em Proceedings of the 49th Annual Meeting of the Association for
  Computational Linguistics: Human Language Technologies-Volume 1\/}.
  Association for Computational Linguistics, pages 142--150.

\bibitem[{Mikolov et~al.(2013)Mikolov, Sutskever, Chen, Corrado, and
  Dean}]{mikolov2013distributed}
Tomas Mikolov, Ilya Sutskever, Kai Chen, Greg~S Corrado, and Jeff Dean. 2013.
\newblock Distributed representations of words and phrases and their
  compositionality.
\newblock In {\em Advances in neural information processing systems\/}. pages
  3111--3119.

\bibitem[{Mnih et~al.(2014)Mnih, Heess, Graves et~al.}]{mnih2014recurrent}
Volodymyr Mnih, Nicolas Heess, Alex Graves, et~al. 2014.
\newblock Recurrent models of visual attention.
\newblock In {\em Advances in neural information processing systems\/}. pages
  2204--2212.

\bibitem[{Nallapati et~al.(2016)Nallapati, Zhou, Gulcehre, Xiang
  et~al.}]{nallapati2016abstractive}
Ramesh Nallapati, Bowen Zhou, Caglar Gulcehre, Bing Xiang, et~al. 2016.
\newblock Abstractive text summarization using sequence-to-sequence {RNN}s and
  beyond.
\newblock In {\em Conference on Computational Natural Language Learning
  (CoNLL)\/}.

\bibitem[{Pang and Lee(2005)}]{pang2005seeing}
Bo~Pang and Lillian Lee. 2005.
\newblock Seeing stars: Exploiting class relationships for sentiment
  categorization with respect to rating scales.
\newblock In {\em Proceedings of the 43rd annual meeting on association for
  computational linguistics\/}. Association for Computational Linguistics,
  pages 115--124.

\bibitem[{Ranzato et~al.(2015)Ranzato, Chopra, Auli, and
  Zaremba}]{RanzatoCAZ15}
Marc'Aurelio Ranzato, Sumit Chopra, Michael Auli, and Wojciech Zaremba. 2015.
\newblock \href{http://arxiv.org/abs/1511.06732}{Sequence level training with
  recurrent neural networks}.
\newblock {\em CoRR\/} abs/1511.06732.
\newblock
  \href{http://arxiv.org/abs/1511.06732}{http://arxiv.org/abs/1511.06732}.

\bibitem[{Rush et~al.(2015)Rush, Chopra, and Weston}]{rush2015neural}
Alexander~M Rush, Sumit Chopra, and Jason Weston. 2015.
\newblock A neural attention model for abstractive sentence summarization.
\newblock In {\em Empirical Methods in Natural Language Processing (EMNLP)\/}.

\bibitem[{Sennrich et~al.(2015)Sennrich, Haddow, and
  Birch}]{sennrich2015neural}
Rico Sennrich, Barry Haddow, and Alexandra Birch. 2015.
\newblock Neural machine translation of rare words with subword units.
\newblock In {\em Annual Meeting of the Association for Computational
  Linguistics (ACL)\/}.

\bibitem[{Seo et~al.(2016)Seo, Kembhavi, Farhadi, and
  Hajishirzi}]{seo2016bidirectional}
Minjoon Seo, Aniruddha Kembhavi, Ali Farhadi, and Hannaneh Hajishirzi. 2016.
\newblock Bidirectional attention flow for machine comprehension.
\newblock {\em arXiv preprint arXiv:1611.01603\/} .

\bibitem[{Sermanet et~al.(2014)Sermanet, Frome, and
  Real}]{sermanet2014attention}
Pierre Sermanet, Andrea Frome, and Esteban Real. 2014.
\newblock Attention for fine-grained categorization.
\newblock {\em arXiv preprint arXiv:1412.7054\/} .

\bibitem[{Shang et~al.(2015)Shang, Lu, and Li}]{shang2015neural}
Lifeng Shang, Zhengdong Lu, and Hang Li. 2015.
\newblock Neural responding machine for short-text conversation.
\newblock In {\em Annual Meeting of the Association for Computational
  Linguistics (ACL)\/}.

\bibitem[{Shen et~al.(2016)Shen, Huang, Gao, and Chen}]{shen2016reasonet}
Yelong Shen, Po-Sen Huang, Jianfeng Gao, and Weizhu Chen. 2016.
\newblock Reasonet: Learning to stop reading in machine comprehension.
\newblock {\em arXiv preprint arXiv:1609.05284\/} .

\bibitem[{Socher et~al.(2011)Socher, Pennington, Huang, Ng, and
  Manning}]{socher2011semi}
Richard Socher, Jeffrey Pennington, Eric~H. Huang, Andrew~Y. Ng, and
  Christopher~D. Manning. 2011.
\newblock Semi-supervised recursive autoencoders for predicting sentiment
  distributions.
\newblock In {\em Proceedings of the conference on empirical methods in natural
  language processing\/}.

\bibitem[{Socher et~al.(2013)Socher, Perelygin, Wu, Chuang, Manning, Ng, Potts
  et~al.}]{socher2013recursive}
Richard Socher, Alex Perelygin, Jean~Y. Wu, Jason Chuang, Christopher~D.
  Manning, Andrew~Y. Ng, Christopher Potts, et~al. 2013.
\newblock Recursive deep models for semantic compositionality over a sentiment
  treebank.
\newblock In {\em EMNLP\/}.

\bibitem[{Sordoni et~al.(2015)Sordoni, Galley, Auli, Brockett, Ji, Mitchell,
  Nie, Gao, and Dolan}]{sordoni2015neural}
Alessandro Sordoni, Michel Galley, Michael Auli, Chris Brockett, Yangfeng Ji,
  Margaret Mitchell, Jian-Yun Nie, Jianfeng Gao, and Bill Dolan. 2015.
\newblock A neural network approach to context-sensitive generation of
  conversational responses.
\newblock {\em arXiv preprint arXiv:1506.06714\/} .

\bibitem[{Sutskever et~al.(2014)Sutskever, Vinyals, and
  Le}]{sutskever2014sequence}
Ilya Sutskever, Oriol Vinyals, and Quoc~V. Le. 2014.
\newblock Sequence to sequence learning with neural networks.
\newblock In {\em Advances in neural information processing systems\/}. pages
  3104--3112.

\bibitem[{Trischler et~al.(2016)Trischler, Ye, Yuan, He, Bachman, and
  Suleman}]{trischler2016parallel}
Adam Trischler, Zheng Ye, Xingdi Yuan, Jing He, Phillip Bachman, and Kaheer
  Suleman. 2016.
\newblock A parallel-hierarchical model for machine comprehension on sparse
  data.
\newblock {\em arXiv preprint arXiv:1603.08884\/} .

\bibitem[{Vinyals and Le(2015)}]{vinyals2015neural}
Oriol Vinyals and Quoc Le. 2015.
\newblock A neural conversational model.
\newblock {\em arXiv preprint arXiv:1506.05869\/} .

\bibitem[{Wang and Jiang(2016)}]{wang2016machine}
Shuohang Wang and Jing Jiang. 2016.
\newblock Machine comprehension using match-lstm and answer pointer.
\newblock {\em arXiv preprint arXiv:1608.07905\/} .

\bibitem[{Wang et~al.(2016)Wang, Mi, Hamza, and Florian}]{wang2016multi}
Zhiguo Wang, Haitao Mi, Wael Hamza, and Radu Florian. 2016.
\newblock Multi-perspective context matching for machine comprehension.
\newblock {\em arXiv preprint arXiv:1612.04211\/} .

\bibitem[{Weston et~al.(2015)Weston, Bordes, Chopra, Rush, van Merri{\"e}nboer,
  Joulin, and Mikolov}]{weston2015towards}
Jason Weston, Antoine Bordes, Sumit Chopra, Alexander~M Rush, Bart van
  Merri{\"e}nboer, Armand Joulin, and Tomas Mikolov. 2015.
\newblock Towards ai-complete question answering: A set of prerequisite toy
  tasks.
\newblock {\em arXiv preprint arXiv:1502.05698\/} .

\bibitem[{Williams(1992)}]{Williams92}
Ronald~J. Williams. 1992.
\newblock Simple statistical gradient-following algorithms for connectionist
  reinforcement learning.
\newblock {\em Machine Learning\/} 8:229--256.

\bibitem[{Wu et~al.(2016)Wu, Schuster, Chen, Le, Norouzi, Macherey, Krikun,
  Cao, Gao, Macherey et~al.}]{wu2016google}
Yonghui Wu, Mike Schuster, Zhifeng Chen, Quoc~V. Le, Mohammad Norouzi, Wolfgang
  Macherey, Maxim Krikun, Yuan Cao, Qin Gao, Klaus Macherey, et~al. 2016.
\newblock Google's neural machine translation system: Bridging the gap between
  human and machine translation.
\newblock {\em arXiv preprint arXiv:1609.08144\/} .

\bibitem[{Xiong et~al.(2016)Xiong, Zhong, and Socher}]{xiong2016dynamic}
Caiming Xiong, Victor Zhong, and Richard Socher. 2016.
\newblock Dynamic coattention networks for question answering.
\newblock {\em arXiv preprint arXiv:1611.01604\/} .

\bibitem[{Zaremba and Sutskever(2015)}]{zaremba2015reinforcement}
Wojciech Zaremba and Ilya Sutskever. 2015.
\newblock Reinforcement learning neural turing machines-revised.
\newblock {\em arXiv preprint arXiv:1505.00521\/} .

\bibitem[{Zhang et~al.(2015)Zhang, Zhao, and LeCun}]{zhang2015character}
Xiang Zhang, Junbo Zhao, and Yann LeCun. 2015.
\newblock Character-level convolutional networks for text classification.
\newblock In {\em Advances in neural information processing systems\/}. pages
  649--657.

\end{thebibliography}
